\documentclass[10pt,twocolumn]{IEEEtran} 

\usepackage{lipsum}
\makeatletter
\def\footnoterule{\relax%
  \kern-5pt
  \hbox to \columnwidth{\hfill\vrule width 1\columnwidth height 0.4pt\hfill}
  \kern4.6pt}
\makeatother

\usepackage{lipsum}

\newcommand\blfootnote[1]{%
  \begingroup
  \renewcommand\thefootnote{}\footnote{#1}%
  \addtocounter{footnote}{-1}%
  \endgroup
}

\usepackage{graphicx}
\usepackage[noadjust]{cite}
 
\usepackage{url}
\def\BibTeX{{\rm B\kern-.05em{\sc i\kern-.025em b}\kern-.08em
    T\kern-.1667em\lower.7ex\hbox{E}\kern-.125emX}}

\begin{document}

\title{An Area and Energy Efficient Design of Domain-Wall Memory-Based Deep Convolutional Neural Networks using Stochastic Computing} 

\author{Xiaolong Ma\textsuperscript{+,1}, Yipeng Zhang\textsuperscript{+,1}, Geng Yuan\textsuperscript{1}, Ao Ren\textsuperscript{1}, Zhe Li\textsuperscript{1},\\ Jie Han\textsuperscript{2}, Jingtong Hu\textsuperscript{3}, Yanzhi Wang\textsuperscript{1}\\
E-mail: \textsuperscript{1}\{xma27, yzhan139, geyuan, aren, zli89, ywang393\}@syr.edu,\\ \textsuperscript{2}jhan8@ualberta.ca, \textsuperscript{3}jthu@pitt.edu}


\maketitle
\thispagestyle{empty}\pagestyle{empty}
\blfootnote{\textsuperscript{+}These authors contributed equally.}

\begin{abstract}
With recent trend of wearable devices and Internet of Things (IoTs), it becomes attractive to develop hardware-based deep convolutional neural networks (DCNNs) for embedded applications, which require low power/energy consumptions and small hardware footprints. Recent works demonstrated that the Stochastic Computing (SC) technique can radically simplify the hardware implementation of arithmetic units and has the potential to satisfy the stringent power requirements in embedded devices. However, in these works, the memory design optimization is neglected for weight storage, which will inevitably result in large hardware cost. Moreover, if conventional volatile SRAM or DRAM cells are utilized for weight storage, the weights need to be re-initialized whenever the DCNN platform is re-started.

In order to overcome these limitations, in this work we adopt an emerging non-volatile Domain-Wall Memory (DWM), which can achieve ultra-high density, to replace SRAM for weight storage in SC-based DCNNs. We propose DW-CNN, the \emph{first} comprehensive design optimization framework of DWM-based weight storage method. We derive the optimal memory type, precision, and organization, as well as whether to store binary or stochastic numbers. We present effective \emph{resource sharing scheme} for DWM-based weight storage in the convolutional and fully-connected layers of SC-based DCNNs to achieve a desirable balance among area, power (energy) consumption, and application-level accuracy.
\end{abstract}


\section{\bf{Introduction}}
  The emerging of autonomous systems, such as unmanned vehicles, robotics, and cognitive wearable devices, imposed a challenge in designing computer systems with machine intelligence. The demand for machine intelligence has been exacerbated by the explosion of the big data, which provides huge potential to enhance business decision making, science discovery, and military or political analysis, etc., albeit whose processing is beyond the capacity of human beings. Recently, deep learning, especially \emph{deep convolutional neural networks} (DCNNs), has been proven to be an effective technique that is capable of handling unstructured data for both supervised and unsupervised learning~\cite{hlee:convolutionalblief,J:deepoverview,ao:learningAI,ding2017c,lin2017fft,wang2018towards,wang2018clstm}. It becomes one of the most promising type of artificial neural networks and has been recognized as the dominant approach for almost all recognition and detection tasks~\cite{bengio:deeplearning}.
  
  The hardware accelerations for DCNNs have been a booming research area on \emph{General-Purpose Graphics Processing Units} (GPGPUs) \cite{GPU_1,GPU_2} and \emph{Field-Programmable Gate Arrays} (FPGAs) \cite{fpga_1,fpga_2,fpga_3,fpga_dadiannao}. Nevertheless, there is a trend of embedding DCNNs into light-weight embedded and portable systems, such as surveillance monitoring systems~\cite{machinglearning_surveillance}, self-driving systems~\cite{selfdriving}, and wearable devices~\cite{wearable}. These scenarios require very low power/energy consumptions and small hardware footprints, and necessitate the investigation of novel hardware computing paradigms.
  
  Recent works~\cite{li2017towards,Ao:DCNN1,Ao:DCNN2,yuan2017softmax,li2017hardware,li2017normalization} considered the \emph{Stochastic Computing} (SC) technique~\cite{SC} as a low-cost substitute to conventional binary-based computing for DCNNs. The SC technique has also been investigated on neural networks and \emph{Deep Belief Networks} (DBNs)~\cite{korean:belief_network,other:belief_network}. SC can radically simplify the hardware implementation of arithmetic units, which are resource-consuming in binary designs, and has the potential to satisfy the low-power requirements of DCNNs. It offers a colossal design space for optimization due to its reduced area and high soft error resiliency. However, the works \cite{Ao:DCNN1,Ao:DCNN2} exhibit certain shortcomings of neglecting the memory design optimization as the memory storage requirements in state-of-the-art DCNNs has become highly demanding \cite{other:weightstorage:required} (especially to store weights in fully-connected layers.) Moreover, if conventional volatile SRAM or DRAM cells are utilized for weight storage, the weights need to be initialized whenever the hardware DCNN platform is powered on, which hurdles the desirable ``plug-and-play'' property of such platforms. 
  
  \begin{figure}[t]
  	\centering
  	\includegraphics[width=3in]{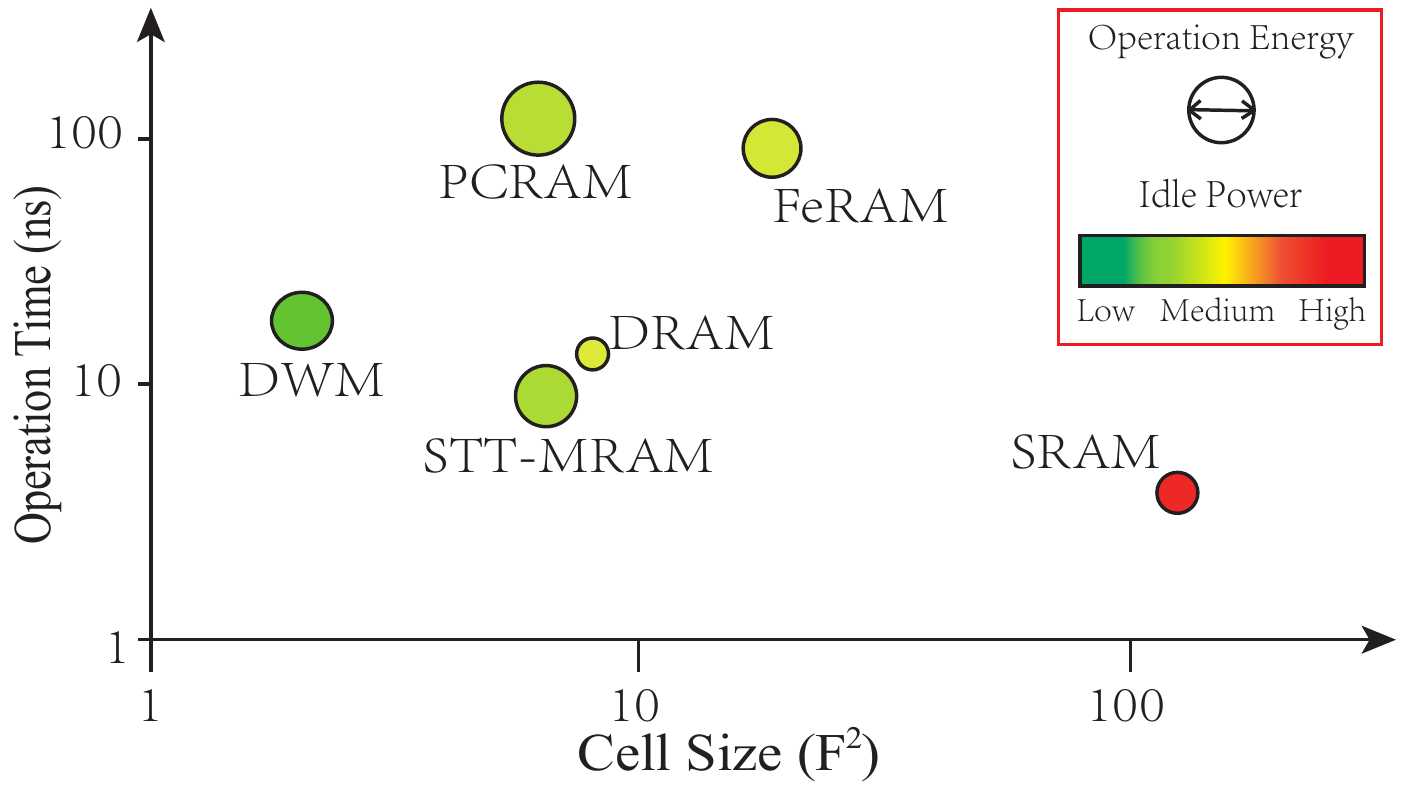}
  	\vskip -0.4em
  	\caption{Comparison of key factors of different memory technologies, (\textit{data from}~\cite{parkin:racetrack})}
  	\label{memory:comp}	
  \end{figure}
  
  Recent breakthroughs in several \emph{non-volatile memory} (NVM) techniques, such as Ferroelectric RAM (FeRAM), Spin-Transfer Torque Magnetic RAM (STT-MRAM), and Domain-Wall Memory (DWM), can potentially replace conventional SRAMs in the neuromorphic systems in order to satisfy the non-volatility and high density/low power requirements. Figure \ref{memory:comp} shows the comparison of various key factors of different memory technologies. The Domain-Wall Memory (DWM), a spintronic non-volatile memory technology, can achieve one of the highest densities (40$\times$ over SRAM) with similar read/write time and idle power while maintain near-zero leakage power compared with others~\cite{spintronic:domainmove,parkin:racetrack,spintronic,tapecache}. In DWM, multiple bits are stored in a nanowire in high storage density, which is suitable for weight storage and retrieval. The unique characteristic of DWM can drastically reduce the neuron size and enlarge the design scalability of DCNNs.
  \begin{figure}[t]
  	\centering
  	\includegraphics[width=3.5in]{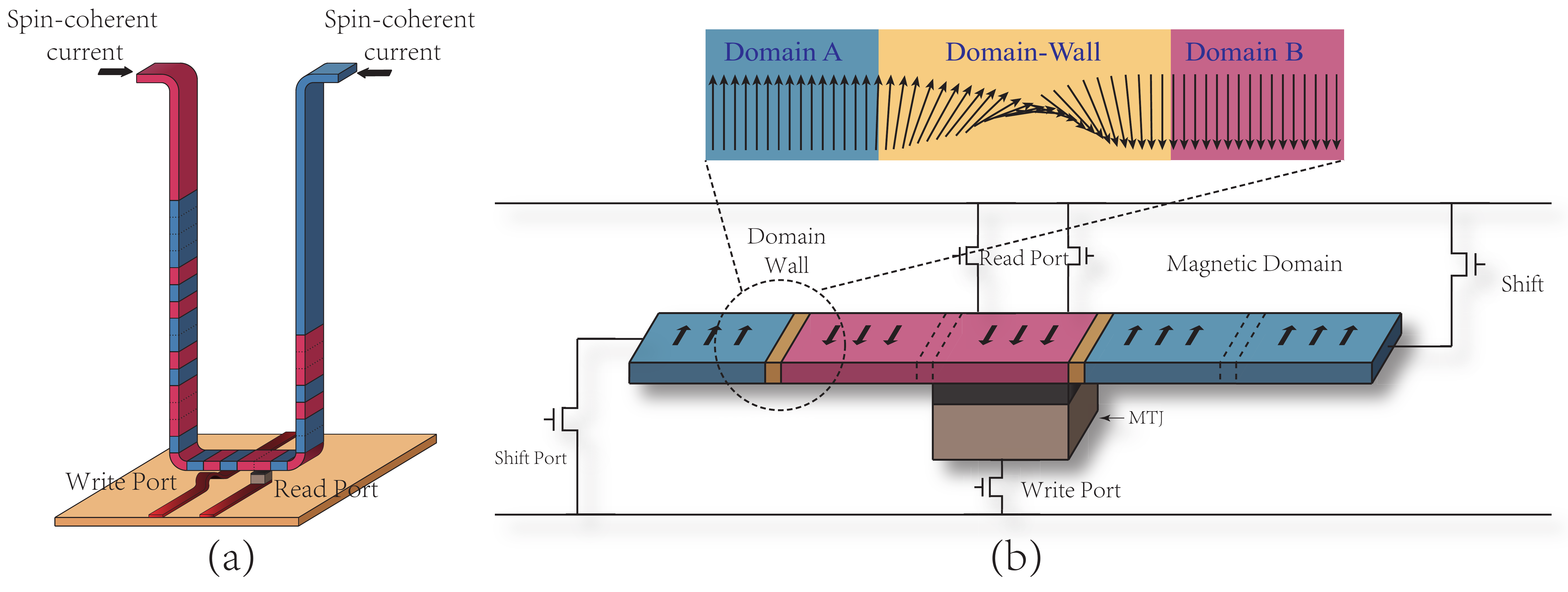}
  	\vskip -0.4em
  	\caption{Structure and operation of the domain-wall memory.}
  	\label{dwm:overview}
  \end{figure}
  
  \begin{figure}[t]
  	\centering
  	\includegraphics[width=3.5in]{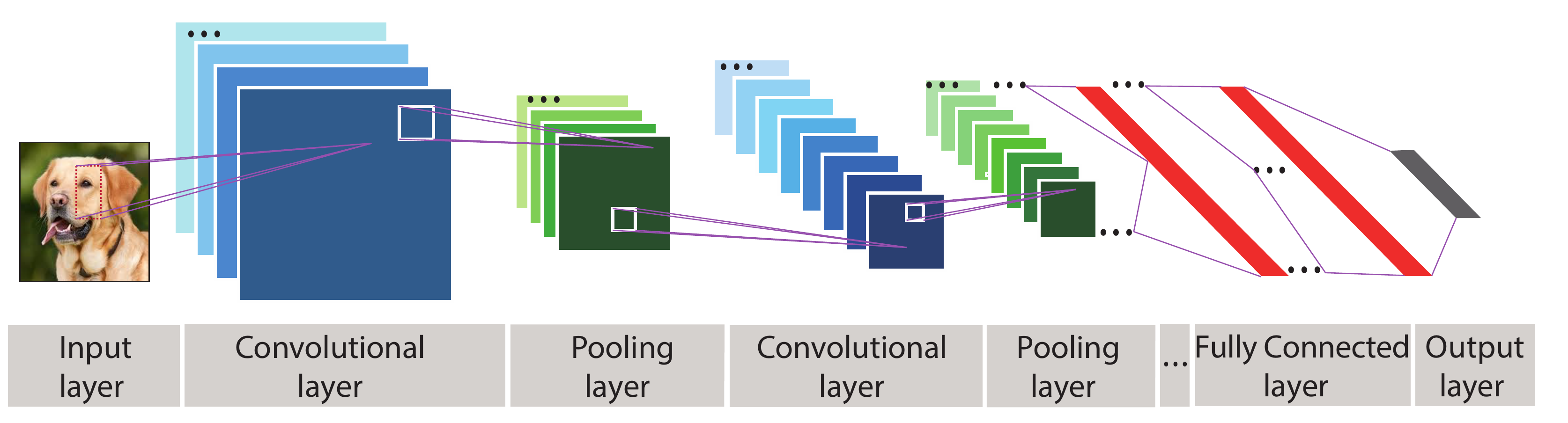}
  	\vskip -0.4em
  	\caption{The general DCNN architecture.}
  	\label{dcnn:overview}
  \end{figure}
  In this paper, we propose DW-CNN, the \emph{first} comprehensive design optimization framework of SC-based DCNNs using domain-wall memory as the weight storage method. We start from a SC-based DCNN system motivated by~\cite{Ao:DCNN1,Ao:DCNN2}, and derive the most efficient weight storage scheme, including the memory type, precision, and organization.  The objective is to reduce area/hardware cost and energy/power consumptions meanwhile maintaining a high application-level accuracy for DCNN. We investigate replacing SRAM by DWM for weight storage and storing binary or stochastic numbers. Besides, we present effective \emph{resource sharing scheme} for DWM-based weight storage in the convolutional and fully-connected layers of SC-based DCNNs, and derive the optimal resource sharing to achieve a desirable balance among area, power (energy) consumptions, and application-level accuracy. Experimental results demonstrate the effectiveness of the proposed DW-CNN framework in area/hardware cost and energy consumption reductions.


%
\begin{figure}[t]
	\centering
	\includegraphics[width=2in]{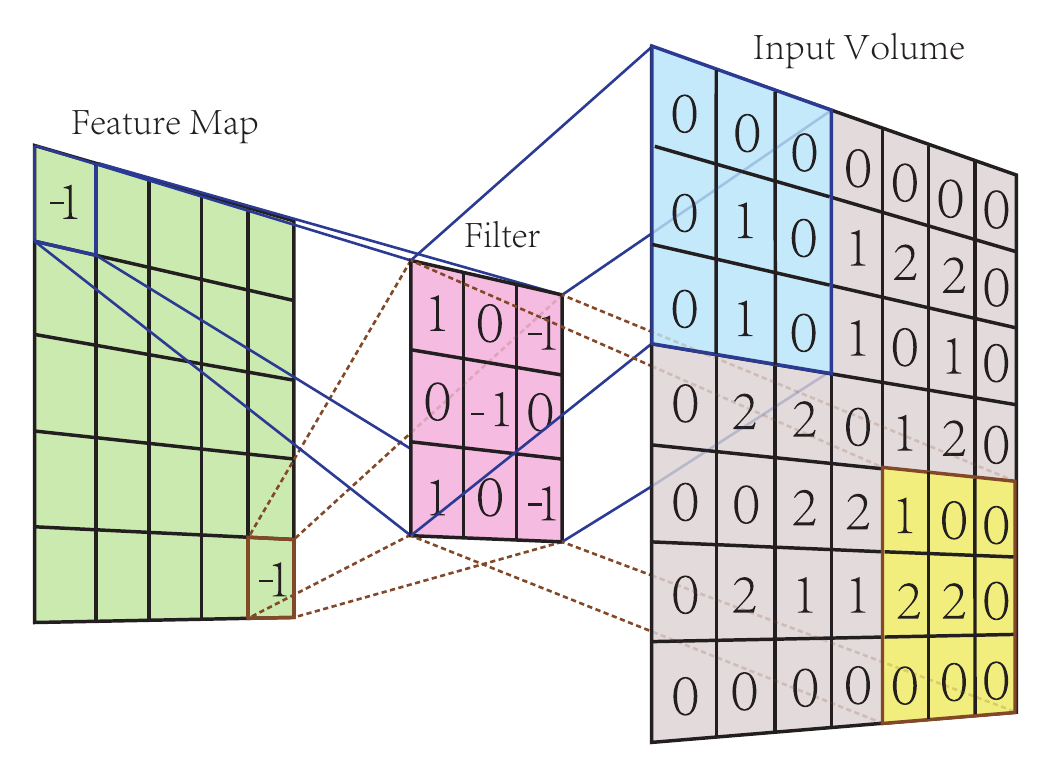}
	\vskip -0.4em
	\caption{Illustration of the convolution process.}
	\label{conv:overview}
\end{figure}

\section{\bf{Overview}}

\subsection{\bf{Domain-Wall Memory Overview}}

Domain-Wall Memory (DWM), a recently developed spin-based non-volatile memory hardware in which multiple bits are stored in a ferromagnetic nanowire which are organized in array of vertical columns on a silicon wafer~\cite{parkin:racetrack}. Figure \ref{dwm:overview}-(a) illustrates the typical structure of a single-wire DWM in 3-D view. As shown in Figure \ref{dwm:overview} (b), the information bits in DWM are separated by magnetic domain walls, and two magnetization directions can represent the binary number 0 or 1. A domain wall is a magnetically neutral zone separating two different polarization domains. Figure \ref{dwm:overview}-(b) illustrates the domain wall as a interface of a gradual re-orientation of the magnetic moments between two 180-degree domains. A spin-coherent electric current, when applying on the shift port at the two ends, can move all domains and domain walls to the left or right at the same velocity without overwriting previous bits~\cite{parkin:racetrack}.

The reading and writing operation in a DWM is shown in Figure \ref{dwm:overview}-(b). Reading data is achieved by measuring the tunnel magneto-resistance of a magnetic tunnel junction (MTJ) unit, which is formed by an insulator separating a strong ferromagnetic layer from the domain wall nanowire~\cite{spintronic}. Moreover, writing data in the domain wall nanowire is accomplished by the fringing field of a domain wall moved in the write port, which can alter the magnetization with a spin-coherent electric current. In addition, the read port (Read Wordline) has one transistor and write port (Write Wordline) has two transistors. Note that the write and read operation can only occur in the MTJ. Therefore, the current needs to ``push" the bit until it is aligned with the fixed layer, while the shift direction and velocity are controlled by the current direction and amplitude~\cite{spintronic:domainmove}.

\subsection{\bf{Overview of DCNN} }

The basic architecture of deep convolutional neural network is inspired by the biological feature of animal visual vertex, which contains two types of cells and are only sensitive to a small region of the visual field~\cite{DCNN:Lenet}. Unlike the traditional fully connected neural networks, a neuron in DCNN is only connected to a small region of the previous layer.

\begin{figure}[t]
			\centering
			\includegraphics[width=3.5in]{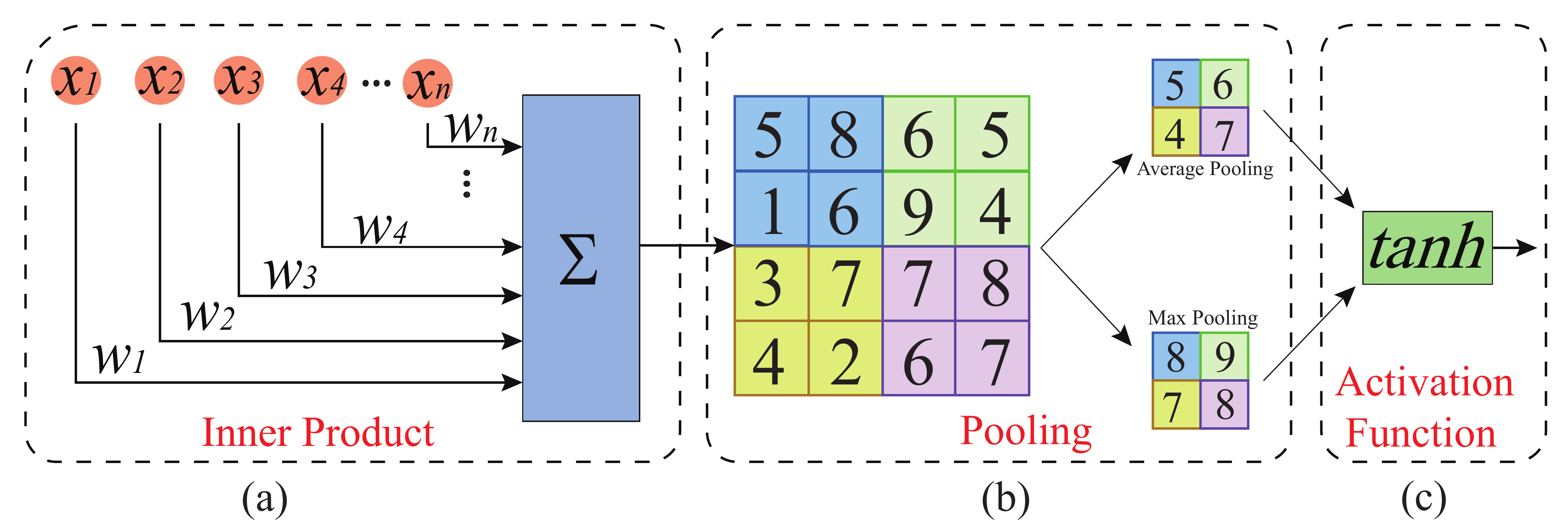}
			\vskip -0.4em
			\caption{Basic operations in DCNN. (a) Inner product, (b) pooling, (c) activation function.}
			\label{dcnn:basicoperation}
\end{figure}

Figure \ref{dcnn:overview} illustrates the widely used DCNN architecture LeNet-5~\cite{LeNet5}.
There are three types of layers in DCNN: {\textit{Convolutional Layer, Pooling Layer,} and {\textit{Fully Connected Layer}. The convolutional layer is a unique block of DCNN, and calculates the inner product of the receptive fields and a set of learnable filters to extract the feature of input~\cite{cs231n}. Figure \ref{conv:overview} illustrates the process of feature extraction by convolution operations. The input feature map size is 7$\times$7, and the filter size is 3$\times$3. Suppose the stride is two, then the result of the convolution will have nine elements.
		
		The outputs of the convolution layer are fed to the pooling layer. There are two common strategies of pooling: {\textit{max pooling}} and {\textit{average pooling}}. Max pooling is to select the maximum value in the selected region, and average pooling is to calculate the average value in the selected region. Pooling process will further reduce the dimension of data. In this paper, we adopt max pooling as our pooling strategy because of the better application-level performance in convergence speed and classification accuracy. After the pooling process, the data will be sent to the {\textit{activation function}}. Different activation functions can apply to DCNNs. However, the most suitable one for stochastic computing applications is the hyperbolic tangent (tanh)~\cite{Larkin:efficienthardware,hinton:imageclassify} because it can be efficiently implemented as a finite state machine with stochastic inputs or an up/down counter with binary inputs.
		
		In a DCNN, the high-level reasoning is done by the fully connected layer which takes outputs from all neurons in previous layer. Based on our experiment, the fully connected layer is the least sensitive to the correlation between weights. So a novel architecture of the fully connected layer with optimized neuron structure will have a great potential to reduce the neuron size as well as power consumption.

		Three main operations of DCNN, i.e., inner product, pooling, and activation function, are shown in Figure \ref{dcnn:basicoperation}, and these operations are cascadedly connected in DCNN. Please note that the inner product operation is used in both convolution and fully-connected neuron, but with different scales.

		\subsection{\bf{Stochastic Computing (SC)}}
		In SC, a stochastic number is utilized to represent a real number by counting the number of 1's in a bit-stream. In the unipolar format, a real number $x$ is represented by a stochastic stream $X$, satisfying $P(X=1)=P(X)=x$. For example, the bit-stream 1100101110 contains six 1's in a ten-bit stream, so it represents a number $P(X=1)=6/10=0.6$. In bipolar format, a real number can be represented by stochastic bit-stream $X$ satisfying $2P(X=1)-1=2P(X)-1=x$, thus 0.6 can be represented by 1101011111. The motivation of using stochastic computing is that it greatly simplifies the involved computing elements, which offers an immense design and optimization space. In this paper, the bipolar format is adopted because the input signals and weights can be both positive and negative. In SC, stochastic bit-streams can be generated efficiently using random number generators (RNGs)~\cite{korean:oldgenerator} or effective RNG sharing methods like the one proposed in \cite{korean:generator}.
		
		\begin{figure}[t]
			\centering
			\includegraphics[width=2.8in]{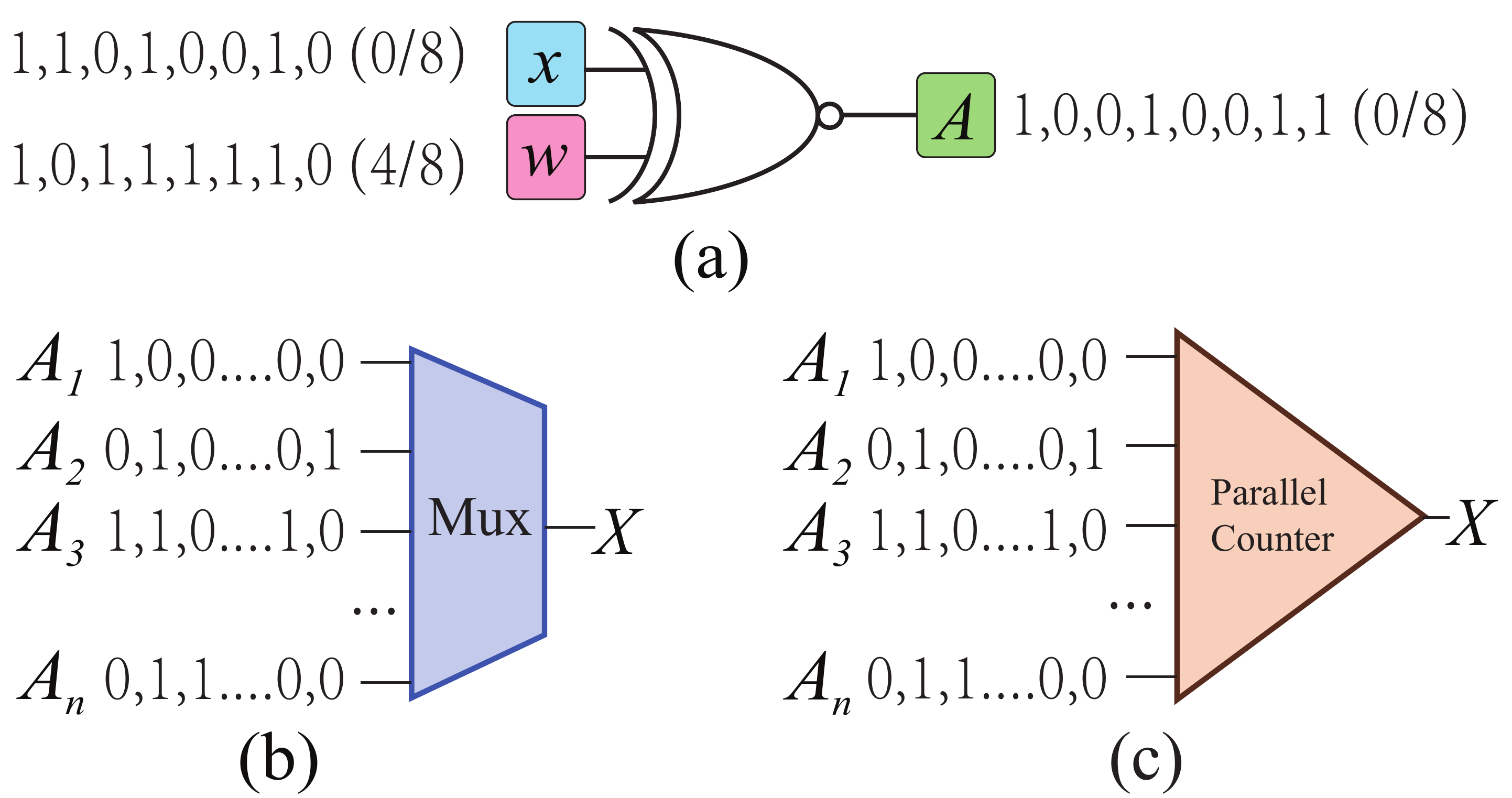}
			\vskip -0.4em
			\caption{SC components: (a) bipolar multiplication, (b) mux-based addition, and (c) APC-based addition.}
			\label{SC:component}
		\end{figure}
		
		$\bullet$ \textbf{\textit{SC Multiplication}}. The multiplication in SC domain can be easily performed by an XNOR gate for the bipolar format. Figure \ref{SC:component}-(a) depicts the bipolar multiplication process of $c=ab$ by XNOR gate which is $c=2P(c=1)-1=2(P(A=1)P(B=1)+P(A=0)P(B=0))-1=(2P(A=1)-1)(2P(B=1)-1)=ab$. 
		
		$\bullet$ \textbf{\textit{SC Addition}}. The objective of addition in SC domain is to calculate the summation of 1's of input stochastic bit-streams. Figure \ref{SC:component}-(b)(c) show two widely used hardware implementations for SC addition: mux-based addition and \emph{approximate parallel counter} (APC)-based addition. For the former structure, a bipolar addition is calculated as $c=2P(C=1)-1=2(\frac{1}{2}P(A=1)+\frac{1}{2}P(B=1))-1=\frac{1}{2}(2P(A=1)-1)+\frac{1}{2}(2P(B=1)-1))=\frac{1}{2}(a+b)$. On the other hand, the APC uses parallel counter to count the total number of 1's among all input bit-streams and outputs a binary number~\cite{Korean:derandomizer}. The mux-based design has a simple structure and is suitable for addition with a small number of inputs, but exhibits inaccuracy when the number of inputs becomes large. The APC-based design is very accurate and is suitable for a large number of inputs, at the expense of more complicated circuit structure. 
		
		\section{\bf{SC-Based DCNN Designs}}
		Motivated by the prior works \cite{Ao:DCNN1,Ao:DCNN2} on SC-based DCNNs, we choose effective SC-based designs for the key operations in DCNNs as summarized in the following (also shown in Figure \ref{neuron:structure}):

		\begin{figure}[h!]
			\centering
			\includegraphics[width=3in]{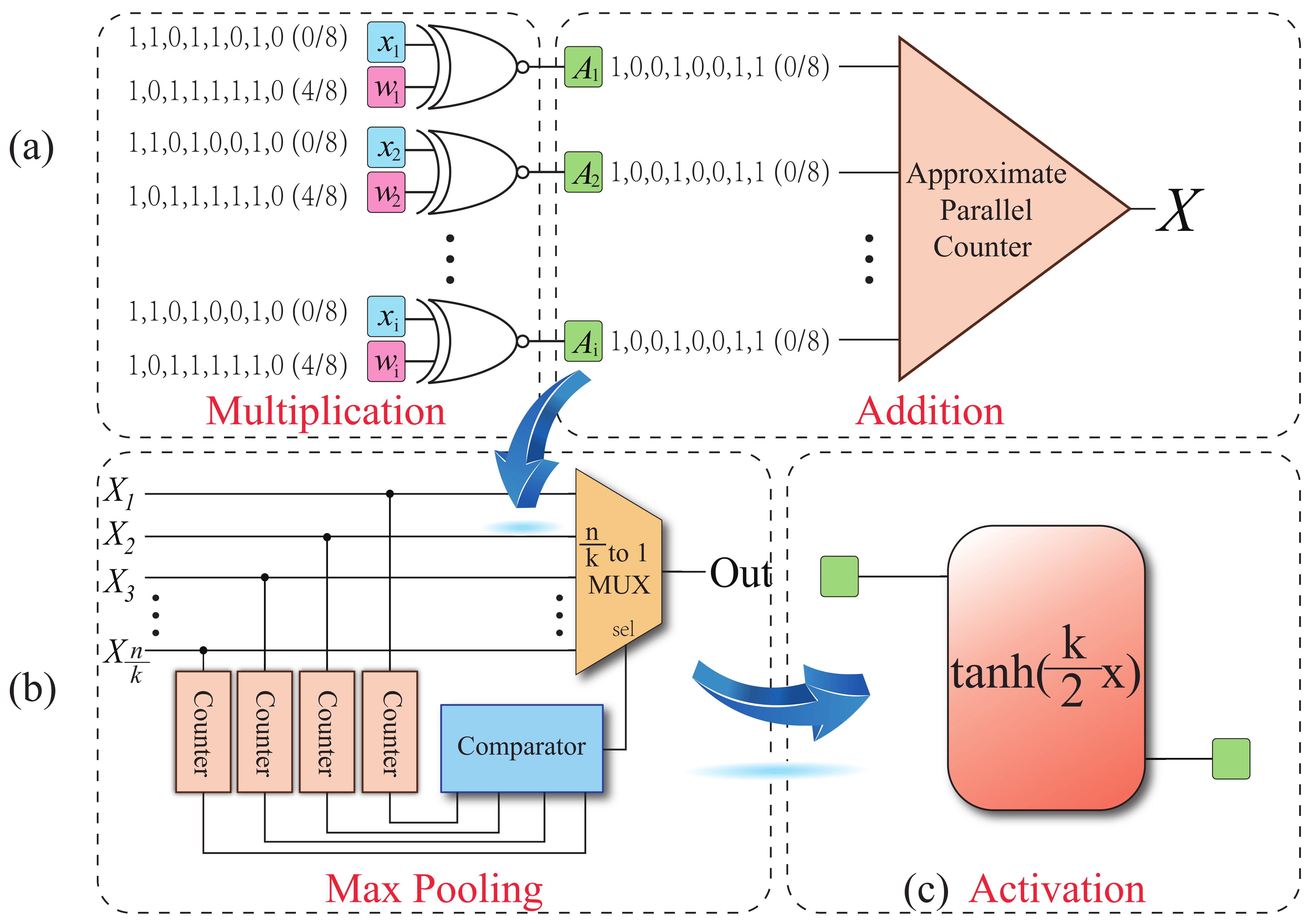}
			\vskip -0.4em
			\caption{SC-based operations in DCNNs: (a) APC-based inner product, (b) SC-based max pooling, and (c) Btanh activation function.}
			\label{neuron:structure}
		\end{figure}
		
		\begin{enumerate}
			\item We choose APC-base addition (together with XNOR-based multiplication) to implement the inner product operation because of the high accuracy when the number of inputs is large, which is the case for neurons in the fully-connected layers. 
			\item We design a hardware-oriented approximation in SC domain for the max pooling operation, which has a light hardware cost and will not incur any extra latency. More specifically, the input bit-streams (inner product results) with length $n$ are divided into bit-segments each with length $k$, and therefore there are $\frac{n}{k}$ segments in each bit-stream. We deploy a segment counter for each input to count the number of 1's in a bit-segment, and a comparator as shown in Figure \ref{neuron:structure} (b). The basic idea is to find the largest number represented by the \emph{current} bit-segment, suppose from input $i$, using segment counters and comparator, and then predict that $i$ is the maximum in the \emph{next} segment. Note that the first segment is randomly chosen in order not to incur any extra latency in clock cycles. This hardware-oriented design successfully avoids the extra latency incurred if max pooling is performed in a straightforward way by counting the number of 1's in all input bit-streams (inner product values).
			\item We adopt the tanh activation function in SC-based DCNN designs because of its simplicity in SC-based implementations and relatively high performance~\cite{Larkin:efficienthardware}. Because the output of APC is a binary number, we adopt the Btanh design which can generate stochastic outputs based on binary inputs~\cite{korean:belief_network}. The Btanh function is implemented using an up/down counter using small hardware footprint.
		\end{enumerate}
		
		\section{\bf{DW-CNN Design Optimization for Weight Storage}}
		In state-of-the-art DCNNs, the requirement of weight storage, in both convolutional and fully-connected layers, is becoming highly demanding. For example, the large-scale AlexNet requires 0.65 million weights for storage~\cite{hinton:imageclassify}. In order to address this challenge, we derive the most efficient weight storage scheme, including the memory type, precision, and organization, for SC-based DCNNs. The objective is to reduce area/hardware cost and energy/power consumptions meanwhile maintaining a high application-level accuracy for the DCNN. We first present a simple but effective \emph{weight reduction method} which is applicable to both SRAM-based or non-volatile memory-based weight storage. Then we will investigate replacing SRAM by DWM and storing binary or stochastic numbers as weights. Finally, we present \emph{resource sharing scheme} for DWM-based weight storage in SC-based DCNNs, and derive the optimal resource sharing to achieve a desirable balance among area, power (energy), and application-level accuracy. 
		
		Overall, in this section we answer the following three questions: (i) What will be the gains when replacing SRAM by DWM in SC-based DCNNs? (ii) Whether it is desirable to store binary number or stochastic number as weights? (iii) What will be the best resource sharing scheme and the corresponding (binary or stochastic) number for storage?
		
		\subsection{\bf{Weight Reduction for SC-Based DCNNs}}
		\begin{figure}[t]
			\centering
			\includegraphics[width=2.5in]{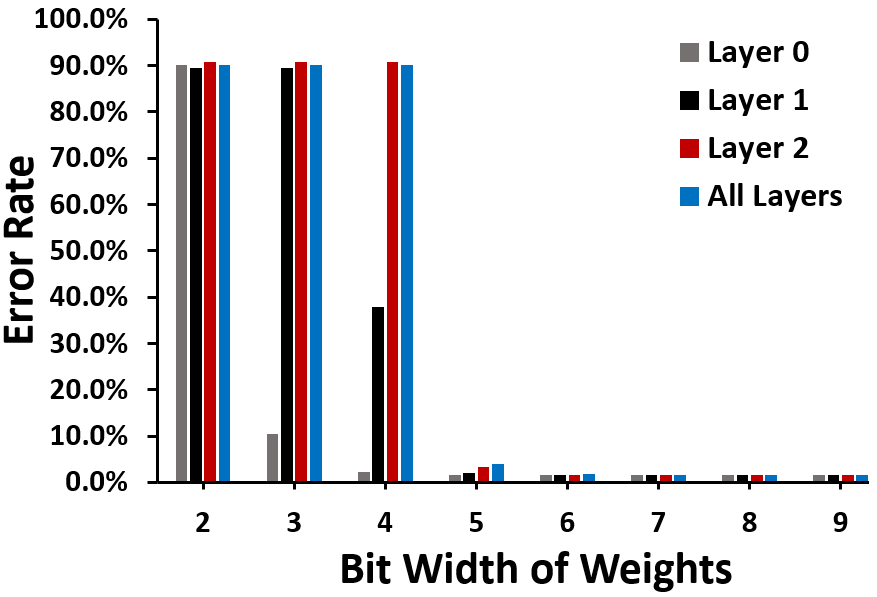}
			\vskip -0.4em
			\caption{Impact of weight precision at each layer on the overall application-level accuracy of LeNet-5 DCNN}
			\label{weight:precision_impact}
		\end{figure}

		By trading off the precision and hardware resources, we can reduce the size and operation energy of the memory for weight storage. According to our software simulation results, many least significant bits which are far from the decimal point can be eliminated without turning into a significant drop in overall accuracy. We adopt a mapping equation that converts a real number to binary number which is stored as the weight value. Suppose the weight value (in real number) is {\textit{x}}, and the number of binary bits to represent weight value is {\textit{w}} (i.e., weight precision), then the binary number {\textit{y}} is:
		\begin{equation} y=\frac{int((x+1)/2\times2^w)}{2^w} \end{equation}
		where {\textit{int}} is the operator of keeping the integer part. In our experiments, we simulated the application-level error rates in terms of different weight precisions. Figure \ref{weight:precision_impact} shows the simulation results of the application-level error rates at different weight precisions on a single layer or all layers.
		Based on our results, the error rates in both single layer and all layers are low enough when the weight precision is seven bits. Thus, with the decrease of weight length, we can implement weight storage more efficiently. Please note that this weight reduction technique is applicable to both SRAM-based and non-volatile memory-based weight storages.
		
		\subsection{\bf{DWM for Weight Storage}}
		SRAM is most often utilized for weight storage in neuromorphic computing systems because of the high reliability and fast reading/writing speed. However, due to the inherent volatility, the weights need to be re-initialized whenever the hardware platform is turned on. Besides, SRAM cells exhibit relatively large size due to the 6-transistor structure, which increases the neuron size and impairs scalability. Both limitations necessitate the investigation of the emerging small-footprint non-volatile memory devices for effective weight storage systems.

		Among emerging non-volatile memory devices, DWM outperforms most of the others in area efficiency. Besides, DWM is especially suitable for storing stochastic numbers because the data are stored in ferromagnetic nanowire ``strips''. We investigate replacing SRAM cells by non-volatile DWM devices for weight storage, and moreover, whether it is desirable to store binary number or stochastic number as weights for SC-based DCNNs.
		
		There are pros and cons on both storing binary numbers and stochastic numbers: Storing binary numbers requires a smaller area/hardware cost for storing the memory bits, but incurs larger area and energy consumption on the peripheral circuits because RNGs and comparators are required to convert binary to stochastic numbers for the inner product computation~\cite{korean:generator}. On the other hand, storing stochastic numbers requires a relatively large area/hardware cost for memory bits storage, but does not need binary-to-stochastic conversion circuitry since the stored stochastic numbers can be directly utilized for computation. Another observation is that when binary number is stored, all bits in the weight need to be retrieved simultaneously to convert to stochastic numbers~\cite{korean:oldgenerator}. On the other hand, when stochastic number is stored, the bits can be read out sequentially, which incurs less energy consumption and is a natural fit to the DWM.
		
		\begin{table}[h!]
			\vskip-0.5em
			\centering
			\caption{Comparison results on the weight storage of LeNet-5 for (i) SRAM vs. DWM and (ii) binary vs. stochastic numbers.}
			\vskip -0.5em
			\resizebox{\columnwidth}{!}{
				\begin{tabular}{c c c c c} 
					\hline
					Weights Data Type & \multicolumn{2}{c}{7-bit Binary number} & \multicolumn{2}{c}{128-bit Stochastic number} \\

					\hline
					Memory Type & SRAM  & DWM  &SRAM  & DWM \\
					Area($mm^2$) & 6.12 & 5.96 & 3.64 & 0.15 \\
					Power($W$) & 1.06 & 0.76 & 7.04 & 0.46 \\
					
					\hline
			\end{tabular} }
			
			\label{why_choose_DWM}
		\end{table}

		We perform testing on the weight storage of LeNet-5 DCNN to compare (i) SRAM and DWM-based weight storage and (ii) storing using binary or stochastic numbers. The comparison results on area and power consumptions are shown in Table \ref{why_choose_DWM}. The overheads of read/write circuitry and binary-to-stochastic conversion circuitry are accounted for. As shown in the table, DWM-based weight storage outperforms SRAM in area and power consumptions due to the high area density and zero leakage power consumptions. Moreover, storing stochastic number in DWM-based weight storage is significantly more desirable because the benefits of avoiding binary-to-stochastic conversion circuitry outweights the larger amount of bits for storage, demonstrating the suitability of DWMs for storing stochastic numbers. Please note that this trend is different for SRAM-based weight storage because of the significantly reduce area consumption of DWM cells.
        
        \begin{figure}[t]
			\centering
			\includegraphics[width=3.5in]{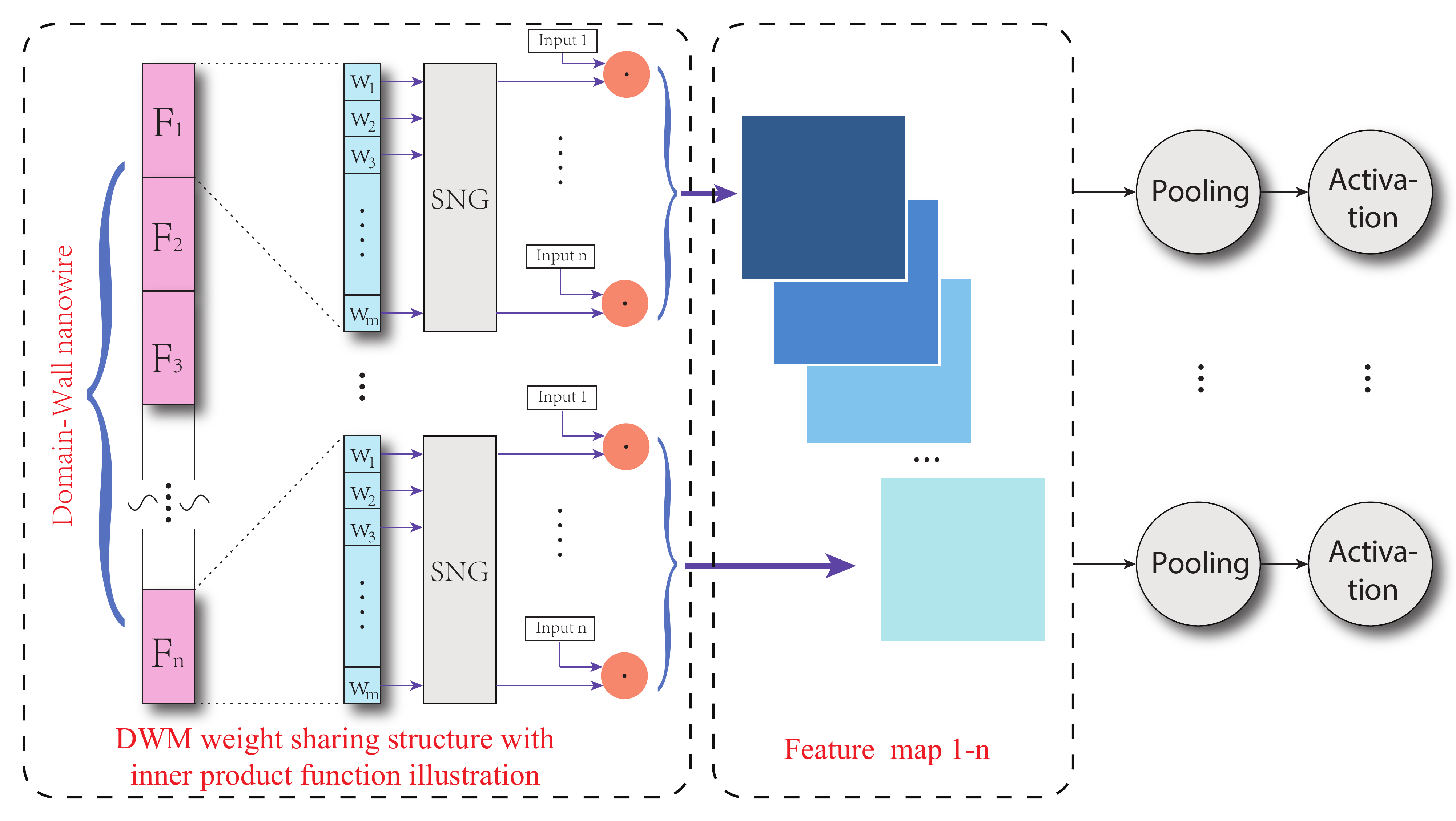}
			\vskip -0.4em
			\caption{Resource (weight) sharing scheme for convolutional layers}
			\label{weightshare01}
		\end{figure}

		\begin{figure*}[htbp]
			\centering
			\includegraphics[width=7.2in]{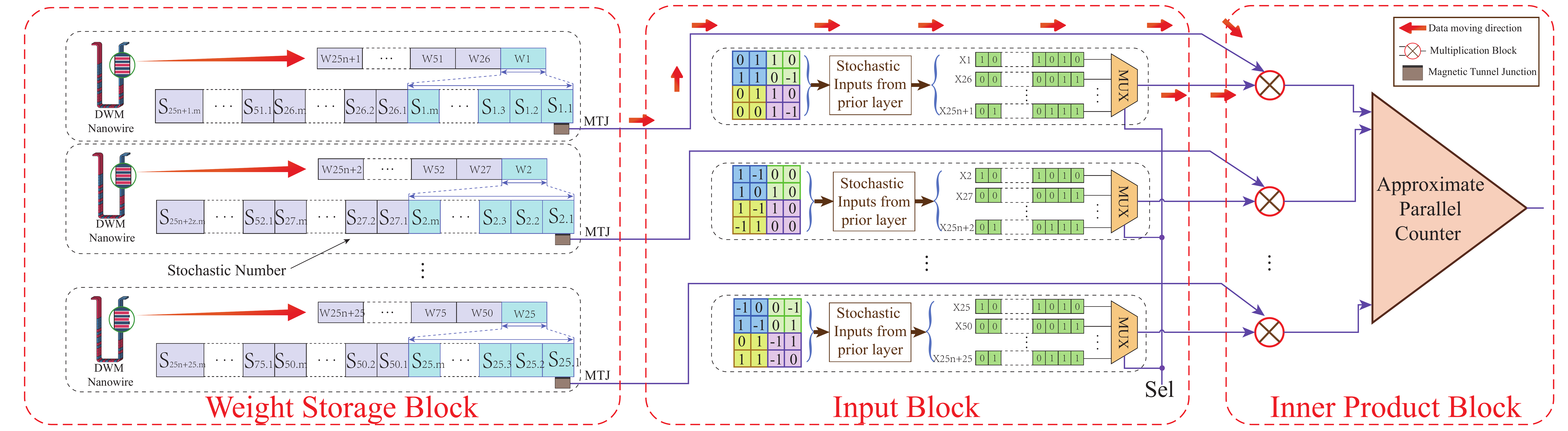}
			\vskip -0.4em
			\caption{Resource (inner product block) sharing scheme for fully-connected layers}
			\vskip -1em
			\label{weighshare2}
		\end{figure*}
		
		\subsection{\bf{Efficient Resource Sharing Schemes on a Layer-wise Consideration}}
		Structural optimizations and resource sharing can further reduce the hardware cost and energy consumptions associated with weight storage in SC-based DCNNs. In this section, we present effective resource sharing methods for DWM-based weight storage in SC-based DCNNs. We present different methods for the convolutional layer and the fully-connected layer fully exploiting the distinct properties of these two types of layers.

		\subsubsection{\bf{Weight Sharing for Convolutional Layers}}
		In order to reduce the amount of hardware resource for filter weight storage in convolutional layers, we develop a weight sharing scheme based on a key observation that the same filter can be applied to all inputs that correspond to one whole feature map \cite{bengio:deeplearning}. In this way, we can separate one DWM ``strip'' into multiple filter-based blocks and each block of filter weights is shared by all inner product blocks for extracting one feature map. The proposed filter weight sharing scheme is illustrated in Figure 9, in which binary-based storage is assumed (and stochastic number is generated by the stochastic number generator (SNG)) without loss of generality. This scheme can significantly reduce the hardware cost for filter weight storage, along with routing overhead and wire delay.

		\subsubsection{\bf{Resource Sharing in Fully-Connected Layers}}
		The resource sharing in fully-connected layers is motivated by two observations: (i) the imprecision in calculation at the fully-connected layers has the least significant impact on overall application-level accuracy of DCNN, and (ii) the bit-stream length (in stochastic number) of input bit-streams (typically 512 or 1024) is higher compared with that of weights (e.g., $2^7=128$ bits as shown in Table 1). Based on these motivations, we present a novel resource sharing scheme for effective sharing of the (APC-based) inner product block and memory reading/writing circuitry, which could achieve significant reduction in hardware cost and energy/power consumptions while maintaining high overall accuracy.
		
		Figure 10 depicts the proposed design of the resource sharing scheme for the fully-connected layer. We store every weight in the stochastic number format to eliminate binary-to-stochastic conversion circuitry, resulting in lower hardware cost and energy/power consumptions as shown in Section 4.2. Let $S_{i,j}$ represent the $j$-th bit of the $i$-th weight in the stochastic number format. Without loss of generality, Figure 10 demonstrates the case in which the weights are stored using 25 DWM nanowires, and the $k$-th nanowire stores weights $w_k$, $w_{25+k}$, $w_{50+k}$, etc. The input bit-streams from the previous layer are grouped into 25 groups and selected by the multiplexers for the (APC-based) inner product calculation. As a result, only a 25-input APC is required, which incurs significant reductions in hardware cost and energy/power consumption compared with the very large APC (e.g., with 800 inputs for LeNet-5) in the original SC-based DCNN system~\cite{LeNet5}. The reading/writing circuitry of the weight storage block can also be reduced, resulting in additional improvement in area/energy efficiencies.
		
		In the proposed design, the input bit-stream length and weight bit-stream length need to be coordinated in order to make sure that the input and weight bit-streams are aligned. Moreover, the multiplexer and selection signals in the proposed design will inevitably incur imprecisions due to information loss. However, as discussed above, the imprecision in calculation at fully-connected layers has minimum impact on the overall application-level accuracy of DCNN. We will show in the experimental results that the inaccuracies incurred by the multiplexer and input selection will have negligible impact on the overall accuracy.
		
				\begin{table*}[t]
			\centering
			\caption{Hardware performance comparison results on the resource sharing scheme of the fully-connected layer with different lengths for weight storage. (SN: Stochastic Number)}
			\renewcommand{\arraystretch}{1}
			\resizebox{\textwidth}{!}{%
				\begin{tabular}{c c c c c c c c c c c} 
					\hline
					Weight Storage & \multicolumn{2}{c}{32-bit SN} & \multicolumn{2}{c}{64-bit SN}& \multicolumn{2}{c}{128-bit SN} & \multicolumn{2}{c}{512-bit SN} & \multicolumn{2}{c}{1024-bit SN}\\ 
					\hline
					Structure & Shared  & Unshared &Shared  & Unshared & Shared  & Unshared & Shared  & Unshared & Shared  & Unshared \\
					\hline
					
					Power($mW$) & 24.67 & 428.97
					& 26.10 & 430.40 & 28.96 & 433.26 &46.13 &450.43 & 69.02 & 473.31 \\
					Energy($nJ$) & 2071.51 & 2196.31 & 2191.63 & 2203.64 & 2431.86 & 2218.29 & 3873.26 & 2306.18 & 5795.14 & 2423.37 \\
					Area($um^2$) & 18410 & 38860 & 18875 & 39324 & 19804 & 40254 & 25379 & 45829 & 32814 & 53263 \\
					\hline
					
			\end{tabular} }
			
			\vskip -1em
			\label{table:results}
		\end{table*}
		
		\begin{table}[t]
			\centering
			\vskip -0.5em
			\caption{Application-level accuracy of the LeNet-5 DCNN vs. length of weight bit-streams at the last fully-connected layer.}\label{tbl_bitstream}
			\resizebox{\columnwidth}{!}{
				\begin{tabular}{ c|c|c|c|c|c|c }
					
					\hline
					bit-stream length  & 32 &64 & 128 & 256 & 512 & 1024 
					\\
					\hline
					accuracy & 99.04\% & 99.04\% & 98.94\% & 98.95\% & 99.00\% & 99.01\% 
					\\
					\hline
				\end{tabular}
			}
		\end{table}
		
		\section{\bf{Optimization Results}}
		
        	\begin{figure}[h!]
			\centering
			\includegraphics[width=3in]{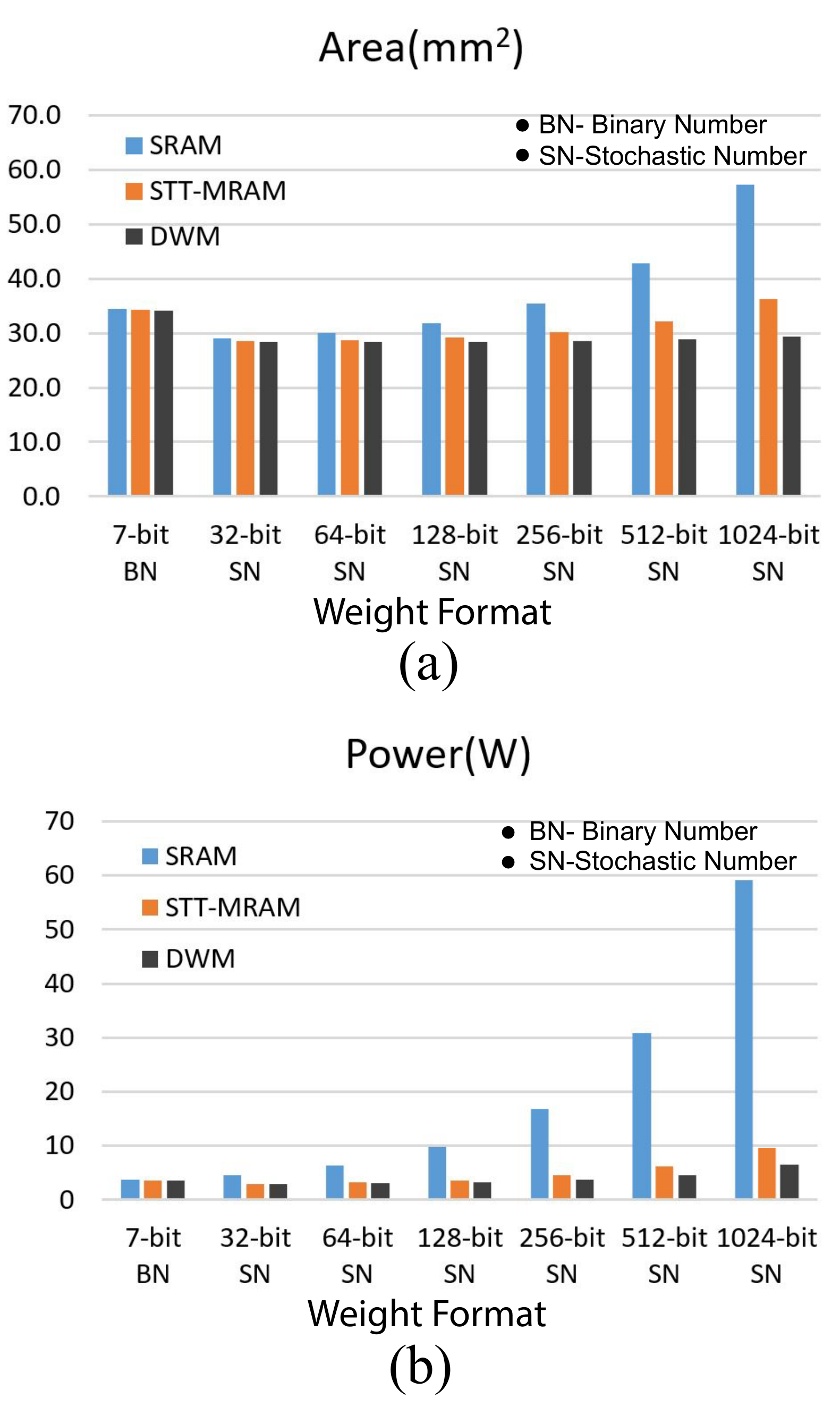}
			\vskip -0.4em
			\caption{Hardware performance on (a) area and (b) power consumption of LeNet-5 DCNN with different memory technologies and different weight storage formats}
			\label{results}
		\end{figure}
		
		We conduct experients on SC-based DCNNs based on the LeNet-5 architecture to reduce area and power consumption, meanwhile keeping a desirable high application-level accuracy. The widely used configuration of LeNet-5 structure is 784-11520-2880-3200-800-500-10, and SC-based DCNNs are evaluated with the MNIST handwritten digit image dataset which contains 60,000 training data and 10,000 testing data. Parameters of key hardware circuitry are obtained by using CACTI~\cite{cacti} for SRAM-based memories and synthesized using Synopsys Design Compiler for logic circuits. Parameters on DWM technology are inhered from~\cite{other:data:ref}, including the DWM reading/writing circuitry.

		First we compare the hardware performance in terms of area and power consumption of the whole LeNet-5 DCNN with different memory technologies and different weight storing formats (binary or stochastic). When the weights are stored in 7-bit binary numbers, the area and power consumption of using different types of memories are almost the same, because the DFFs and SNGs in the network dominate under this scenario. However, when the weights are stored in stochastic numbers and stored with DWMs, the area and power consumption reduce compared with the binary-based cases. The amounts of reductions are less compared with Table 1 because the SC-based computation blocks, e.g., inner product, pooling, activation function, are accounted for in Figure 11 and remain unchanged. Moreover, benefiting from the highly compact cell size of DWMs and high capacity, the area almost does not increase with the increase of stored bits. As for the power consumption shown in Figure \ref{results}-(b), the power consumption using SRAMs dramatically increases with the increasing of storing bits, but the power consumption of using DWMs and stochastic numbers only increases very insignificantly.

		As explained in the previous sections, the calculation imprecisions in the (last) fully-connected layers have relatively insignificant impact on the overall application-level accuracy. For testing, we test different lengths of weight bit-streams (using stochastic numbers) in the last fully-connected layer from 32 to 1024 (assuming an input bit-stream length of 1024) and the overall application-level accuracy of the LeNet-5 DCNN structure. Table 2 illustrates the testing results, which validate the observation as the motivation of our resource sharing scheme. Similarly, a length of weight bit-stream of 256 will yield a high-enough overall accuracy when applied to all the fully-connected layers of LeNet-5.

		Finally, we conduct experiments to test the hardware performance, including power, energy, and area, on the resource sharing scheme of the fully-connected layer using DWM for weight storage. Table 3 provides the results of the whole fully-connected layer at different lengths for weight storage. It can be observed that the resource sharing scheme can reduce the area by up to 52.6\% and reduce the power consumption by 17.35$\times$ compared with the case without resource sharing. The main results of such gains in area and power/energy efficiencies are due to the smaller size of APC (and APC-based inner product block) as well as the sharing of DWM reading/writing circuitry. 

\section{\bf{Conclusion}}

In this paper, we adopt a novel technology of non-volatile Domain-Wall Memory (DWM), which can achieve ultra-high density, to replace SRAM for weight storage in SC-based DCNNs. We proposed the first comprehensive architecture and optimization framework of DW-CNN by developing an optimal scheme of memory type, precision, and organization, as well as whether to store binary or stochastic numbers. We achieve a desirable small size and energy efficient SC-based DCNN while maintaining a very high application-level accuracy.

\section{Acknowledgement}
This work is supported by the National Science Foundation funding awards CNS-1739748 and CNS-1704662.

\bibliographystyle{unsrt}
\bibliography{sigproc}

\end{document}